\documentclass[11pt,letterpaper]{article}
\usepackage{emnlp2017}
\usepackage{times}
\usepackage{latexsym}

\usepackage[inline]{enumitem}
\usepackage{graphicx}
\usepackage{float}
\usepackage{amsmath}
\usepackage{amssymb}
\usepackage{textcomp}
\usepackage{array} 
\usepackage{multirow}

\newcommand{\chapquote}[2]{\begin{quotation} \noindent {\itshape#1} \hfill \text{#2}\end{quotation}  }

\newcolumntype{L}{>{\arraybackslash}m{.95\columnwidth}}
\usepackage[font=normalsize,skip=5pt]{caption}

\emnlpfinalcopy

\def\emnlppaperid{455}

\title{Satirical News Detection and Analysis using Attention Mechanism and Linguistic Features}

\author{Fan Yang \and Arjun Mukherjee \\
        Address line \\ ... \\ Address line}
\author{
        Fan Yang \and Arjun Mukherjee \\ Department of Computer Science \\  University of Houston \\ {\tt \{fyang11,arjun\}@uh.edu}
        \And
        Eduard Gragut \\ Computer and Information Sciences \\ Temple University \\ {\tt edragut@temple.edu}
        }

\date{}

\begin{document}

\maketitle

\begin{abstract}
Satirical news is considered to be entertainment, but it is potentially deceptive and harmful. Despite the embedded genre in the article, not everyone can recognize the satirical cues and therefore believe the news as true news. We observe that satirical cues are often reflected in certain paragraphs rather than the whole document. Existing works only consider document-level features to detect the satire, which could be limited. We consider paragraph-level linguistic features to unveil the satire by incorporating neural network and attention mechanism. We investigate the difference between paragraph-level features and document-level features, and analyze them on a large satirical news dataset. The evaluation shows that the proposed model detects satirical news effectively and reveals what features are important at which level.  
\end{abstract}

\section{Introduction}

\chapquote{``When information is cheap, attention becomes expensive."}{--- James Gleick}

\noindent Satirical news is considered to be entertainment. However, it is not easy to recognize the satire if the satirical cues are too subtle to be unmasked and the reader lacks the contextual or cultural background. The example illustrated in Table~\ref{table:1} is a piece of satirical news with subtle satirical cues. 

\begin{table}[ht]
\small
\begin{tabular}{m{.9\columnwidth}}
\hline
\hline
...\\
``Kids these days are done with stories where things happen," said CBC consultant and world\textquotesingle s oldest child psychologist Obadiah Sugarman. ``We\textquotesingle ll finally be giving them the stiff Victorian morality that I assume is in vogue. Not to mention, doing a period piece is a great way to make sure white people are adequately represented on television." \\
...\\
\hline
\hline
\end{tabular}
\caption{A paragraph of satirical news}
\label{table:1}
\end{table}

Assuming readers interpret satirical news as true news, there is not much difference between satirical news and fake news in terms of the consequence, which may hurt the credibility of the media and the trust in the society. In fact, it is reported in {\it the Guardian} that people may believe satirical news and spread them to the public regardless of the ridiculous content\footnote{\href{https://www.theguardian.com/media/2016/nov/17/facebook-fake-news-satire}{https://www.theguardian.com/media/2016/nov/17/facebook-fake-news-satire}}. It is also concluded that fake news is similar to satirical news via a thorough comparison among true news, fake news, and satirical news~\cite{horne2017just}. This paper focuses on {\it satirical news detection} to ensure the trustworthiness of online news and prevent the spreading of potential misleading information.

Some works tackling fake news and misleading information favor to discover the truth~\cite{xiao2016towards,wan2016truth} through knowledge base~\cite{dong2015knowledge} and truthfulness estimation~\cite{ge2013multi}. These approaches may not be feasible for satirical news because there is no ground-truth in the stories. Another track of works analyze social network activities~\cite{zhao2015enquiring} to evaluate the spreading information~\cite{gupta:2012evaluating,castillo:2011information}. This could be ineffective for both fake news and satirical news because once they are distributed on the social network, the damage has been done. Finally, works evaluating culture difference~\cite{perez2014cross}, psycholinguistic features~\cite{ott:2011finding}, and writing styles~\cite{feng:2012syntactic} for deception detection are suitable for satirical news detection. These works consider features at document level, while we observe that satirical cues are usually located in certain paragraphs rather than the whole document. This indicates that many document level features may be superfluous and less effective.

To understand how paragraph-level features and document-level features are varied towards detection decision when only document level labels are available, we propose a 4-level neural network in a character-word-paragraph-document hierarchy and utilize attention mechanism~\cite{bahdanau:2014neural} to reveal their relative difference. We apply psycholinguistic features, writing stylistic features, structural features, and readability features to understand satire. The paragraph-level features are embedded into attention mechanism for selecting highly attended paragraphs, and the document-level features are incorporated for the final classification. This is the first work that unveils satirical cues between paragraph-level and document-level through neural networks to our knowledge.

We make the following contributions in our paper:
\begin{itemize}
\vspace{-5pt}
    \item We propose a 4-level hierarchical network for satirical news detection. The model detects satirical news effectively and incorporates attention mechanism to reveal paragraph-level satirical cues.
    \item We show that paragraph-level features are more important than document-level features in terms of the psycholinguistic feature, writing stylistic feature, and structural feature, while the readability feature is more important at the document level.
    \item We collect satirical news (16,000+) and true news (160,000+) from various sources and conduct extensive experiments on this corpus\footnote{\href{https://github.com/fYYw/satire}{https://github.com/fYYw/satire}}. 
\end{itemize}

\section{Related Work}
We categorize related works into four categories: content-based detection for news genre, truth verification and truthfulness evaluation, deception detection, and identification of highly attended component using attention mechanism. 

{\bf Content-based detection for news genre}.Content-based methods are considerably effective to prevent satirical news from being recognized as true news and spreading through social media.~\newcite{burfoot:2009automatic} introduce headline features, profanity, and slang to embody satirical news. They consider absurdity as the major device in satirical news and model this feature by comparing entity combination in a given document with Google query results.~\newcite{rubin2016fake} also consider absurdity but model it through unexpected new name entities. They introduce additional features including humor, grammar, negative affect, and punctuation to empower the detection. Besides satirical news,~\newcite{chen:2015misleading} aim to detect click-baits, whose content exaggerates fact.~\newcite{potthast2017stylometric} report a writing style analysis of hyperpartisan news.~\newcite{barbieri2015we} focus on multilingual tweets that advertise satirical news.

It is noteworthy that satirical news used for evaluation in above works are of limited quantity (around 200 articles). Diverse examples of satire may not be included as discussed by~\newcite{rubin2016fake}. This issue inspires us to collect more than 16,000 satirical news for our experiment.

{\bf Truth discovery and truthfulness evaluation}. Although truth extraction from inconsistent sources~\cite{ge2013multi,wan2016truth,li2016verification} and from conflicting sources~\cite{yin2008truth,li2014resolving}, truth inference through knowledge base~\cite{dong2015knowledge}, and discovering evolving truth~\cite{li2015discovery} could help identify fact and detect fake news, they cannot favor much for satirical news as the story is entirely made up and the ground-truth is hardly found. Analyzing user activities~\cite{farajtabar2017fake} and interactions~\cite{castillo:2011information,mukherjee:2015leveraging} to evaluate the credibility may not be appropriate for satirical news as it cannot prevent the spreading. Therefore, we utilize content-based features, including psycholinguistic features, writing stylistic features, structural features, and readability features, to address satirical news detection.   

{\bf Deception detection}. We believe satirical news and opinion spam share similar characteristics of writing fictitious and deceptive content, which can be identified via a psycholinguistic consideration~\cite{mihalcea2009lie,ott:2011finding}. Beyond that, both syntactic stylometry~\cite{feng:2012syntactic} and behavioral features~\cite{mukherjee:2013yelp} are effective for detecting deceptive reviews, while stylistic features are practical to deal with obfuscating and imitating writings~\cite{afroz:2012detecting}. However, deceptive content varies among paragraphs in the same document, and so does satire. We focus on devising and evaluating paragraph-level features to reveal the satire in this work. We compare them with features at the document level, so we are able to tell what features are important at which level. 

{\bf Identification of highly attended component using attention mechanism}. Attention mechanism is widely applied in machine translation~\cite{bahdanau:2014neural}, language inference~\cite{rocktaschel2015reasoning}, and question answering~\cite{chen:2016thorough}. In addition,~\newcite{yang:2016hierarchical} propose hierarchical attention network to understand both attended words and sentences for sentiment classification.~\newcite{chen2016neural} enhance the attention with the support of user preference and product information to comprehend how user and product affect sentiment ratings. Due to the capability of attention mechanism, we employ the same strategy to show attended component for satirical news. Different from above works, we further evaluate linguistic features of highly attended paragraphs to analyze characteristics of satirical news, which has not been explored to our knowledge.

\section{The Proposed Model}
We first present our 4-level hierarchical neural network and explain how linguistic features can be embedded in the network to reveal the difference between paragraph level and document level. Then we describe the linguistic features. 

\subsection{The 4-Level Hierarchical Model}
\begin{figure}[ht]
\centering
\includegraphics[width=.9\columnwidth]{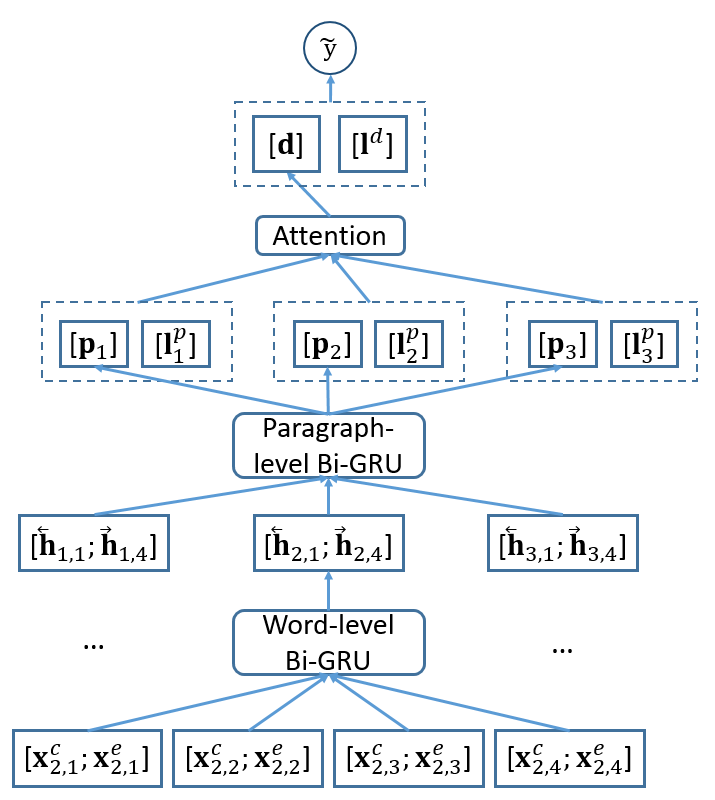}
\caption{The overview of the proposed model. The document has 3 paragraphs and each paragraph contains 4 words. We omit character-level convolution neural network but leave~\(\mathbf{x}^c\) to symbolize the representation learned from it.}
\label{fig:1}
\end{figure}

We build the model in a hierarchy of character-word-paragraph-document. The general overview of the model can be viewed in Figure~\ref{fig:1} and the notations are listed in Table~\ref{table:2}.

\begin{table}[ht]
\centering
\small
\begin{tabular}[width=\columnwidth]{c|l}
\hline
& {\small Meaning} \\
\hline
\multirow{2}{*}{\small Superscript} & Lowercase for notation purpose; \\
& \(\top\) means matrix transpose. \\
\hline
{\small Subscript} & For index purpose. \\
\hline
\multirow{2}{*}{\small Parameter} & \(\mathbf{W}\),\(\mathbf{U}\),\(\mathbf{w}^c\),\(\mathbf{v}^a\): learnable weights; \\
 & \(b\): learnable bias. \\
 \hline
\multirow{5}{*}{\small Representation} & 
\(\mathbf{c}\): character; \(\mathbf{x}\): word; \(\mathbf{p}\): paragraph; \\
 & \(\mathbf{d}\): document;  \(\tilde{y}\): prediction\\ 
 & \(\mathbf{l}\): linguistic vector; \(y\): label;\\
 & \(\mathbf{r}\): reset gate; \(\mathbf{z}\): update gate; \\
 & \(\mathbf{h}\): hidden state for GRU; \\
 & \(\mathbf{u}\): hidden state for attention. \\
\hline
\end{tabular}
\caption{Notations and meanings}
\label{table:2}
\end{table}

\subsubsection{Character-Level Encoder}
We use convolutional neural networks (CNN) to encode word representation from characters. CNN is effective in extracting morphological information and name entities~\cite{ma2016end}, both of which are common in news. Each word is presented as a sequence of~\(n\) characters and each character is embedded into a low-dimension vector. The sequence of characters~\(\mathbf{c}\) is brought to the network. A convolution operation with a filter~\(\mathbf{w}^c\) is applied and moved along the sequence. Max pooling is performed to select the most important feature generated by the previous operation. The word representation~\(\mathbf{x}^c\in\mathbb{R}^f\) is generated with~\(f\) filters.

\subsubsection{Word-Level Encoder}
Assume a sequence of words of paragraph~\(i\) arrives at time~\(t\). The current word representation~\(\mathbf{x}_{i,t}\) concatenates~\(\mathbf{x}^c_{i,t}\) from character level with pretrained word embedding~\(\mathbf{x}^e_{i,t}\), as~\(\mathbf{x}_{i,t}=[\mathbf{x}^c_{i,t};\mathbf{x}^e_{i,t}]\). Examples are given in Figure~\ref{fig:1}. We implement Gated Recurrent Unit (GRU)~\cite{cho2014learning} rather than LSTM~\cite{hochreiter:1997long} to encode the sequence because GRU has fewer parameters. The GRU adopts reset gate \(\mathbf{r}_{i,t}\) and update gate \(\mathbf{z}_{i,t}\) to control the information flow between the input \(\mathbf{x}_{i,t}\) and the candidate state~\(\tilde{\mathbf{h}}_{i,t}\). The output hidden state~\(\mathbf{h}_{i,t}\) is computed by manipulating previous state~\(\mathbf{h}_{i,t-1}\) and the candidate state~\(\tilde{\mathbf{h}}_{i,t}\) regarding to~\(\mathbf{z}_{i,t}\) as in Equation~\ref{eq:4}, where~\(\odot\) denotes element-wise multiplication.

\begin{align}
    \begin{split}
    \label{eq:1}
        \mathbf{z}_{i,t}=\sigma(\mathbf{W}^z\mathbf{x}_{i,t} + \mathbf{U}^z\mathbf{h}_{i,t-1} + b^z)
    \end{split}
\\
    \begin{split}
    \label{eq:2}
        \mathbf{r}_{i,t}=\sigma(\mathbf{W}^r\mathbf{x}_{i,t} + \mathbf{U}^r\mathbf{h}_{i,t-1} + b^r)
    \end{split}
\\
    \begin{split}
    \label{eq:3}
        \tilde{\mathbf{h}}_{i,t} = \text{tanh}(\mathbf{W}^h\mathbf{x}_{i,t} + \mathbf{r}_{i,t}\odot (\mathbf{U}^h\mathbf{h}_{i,t-1}+b^h))
    \end{split}
\\
    \begin{split}
    \label{eq:4}
        \mathbf{h}_{i,t} = (1-\mathbf{z}_{i,t})\odot\mathbf{h}_{i,t-1} + \mathbf{z}_{i,t}\odot\tilde{\mathbf{h}}_{i,t}
    \end{split}
\end{align}

To learn a better representation from the past and the future, we use bidirectional-GRU (Bi-GRU) to read the sequence of words with forward~\(\overrightarrow{\text{GRU}}\) from~\(\mathbf{x}_{i,1}\) to~\(\mathbf{x}_{i,t}\), and backward~\(\overleftarrow{\text{GRU}}\) from~\(\mathbf{x}_{i,t}\) to~\(\mathbf{x}_{i,1}\). The final output of Bi-GRU concatenates the last state of \(\overrightarrow{\text{GRU}}\) and \(\overleftarrow{\text{GRU}}\), as~\([\overrightarrow{\mathbf{h}}_{i,t};\overleftarrow{\mathbf{h}}_{i,1}]\), to represent the \(i\)th paragraph.

\subsubsection{Paragraph-Level Attention}
We observe that not all paragraphs have satire and some of them are functional to make the article complete, so we incorporate attention mechanism to reveal which paragraphs contribute to decision making. Assuming a sequence of paragraph representations have been constructed from lower levels, another Bi-GRU is used to encode these representations to a series of new states~\(\mathbf{p}_{1:t}\), so the sequential orders are considered. 

To decide how paragraphs should be attended, we calculate satirical degree~\(\alpha_i\) of paragraph~\(i\). We first convey~\(\mathbf{p}_i\) into hidden states~\(\mathbf{u}_i\) as in Equation~\ref{eq:5}. Then we product~\(\mathbf{u}_i\) with a learnable satire-aware vector~\(\mathbf{v}^a\) and feed the result into softmax function as in Equation~\ref{eq:6}. The final document representation~\(\mathbf{d}\) is computed as a weighted sum of~\(\alpha_i\) and~\(\mathbf{p}_i\).

\begin{align}
    \begin{split}
    \label{eq:5}
        \mathbf{u}_i = \text{tanh}(\mathbf{W}^a\mathbf{p}_i + b^a)
    \end{split}
\\
    \begin{split}
    \label{eq:6}
        \alpha_i = \frac{\text{exp}(\mathbf{u}_i^\top\mathbf{v}^a)}{\sum_{j=0}^t \text{exp}(\mathbf{u}_j^\top\mathbf{v}^a))}
    \end{split}
\\
    \begin{split}
    \label{eq:7}
        \mathbf{d} = \sum^t_{i=0}\alpha_i\mathbf{p}_i
    \end{split}
\end{align}

Linguistic features are leveraged to support attending satire paragraph. Besides~\(\mathbf{p}_i\), we represent paragraph~\(i\) based on our linguistic feature set and transform it into a high-level feature vector~\(\mathbf{l}_i^p\) via multilayer perceptron (MLP). So~\(\mathbf{u}_i\) in Equation~\ref{eq:5} is updated to: 

\begin{equation}
\label{eq:8}
    \mathbf{u}_i = \text{tanh}(\mathbf{W}^a\mathbf{p}_i + \mathbf{U}^a\mathbf{l}^p_i + b^a)
\end{equation}

\subsubsection{Document-Level Classification}
Similar to the paragraph level, we represent document~\(j\) based on our linguistic feature set and transform it into a high-level feature vector~\(\mathbf{l}^d_j\) via MLP. We concatenate~\(\mathbf{d}_j\) and~\(\mathbf{l}^d_j\) together for classification. Suppose~\(y_j\in(0,1)\) is the label of the document~\(j\), the prediction \(\tilde{y}_j\) and the loss function~\(\mathcal{L}\) over \(N\) documents are:

\begin{align}
    \begin{split}
    \label{eq:9}
        \tilde{y}_j = \text{sigmoid}(\mathbf{W}^d\mathbf{d}_j + \mathbf{U}^d\mathbf{l}^d_j + b^d)
    \end{split}
\\
    \begin{split}
    \label{eq:10}
        \mathcal{L} = -\frac{1}{N}\sum^N_j y_j\log\tilde{y}_j + (1-y_j)\log(1-\tilde{y}_j)
    \end{split}
\end{align}

\subsection{Linguistic Features}

\begin{table*}[ht]
\centering
\small
\begin{tabular}[width=\textwidth]{c|*{8}{|c|}|c}
\hline
& {\small \#Train} & {\small \#Validation} & {\small \#Test} & {\small \#Para} & {\small \#Sent} & {\small \#Words} & {\small \# Capitals} & {\small \#Punc} & {\small \#Digits}\\
\hline
True & 101,268 & 33,756 & 33,756 & 20\(\pm\)7.8 & 32\(\pm\)24 & 734\(\pm\)301 & 118\(\pm\)58 & 28\(\pm\)26 & 93\(\pm\)49 \\
\hline
Satire & 9,538 & 3,103 & 3,608 & 12\(\pm\)4.4 & 25\(\pm\)12 & 587\(\pm\)246 & 87\(\pm\)44 & 11\(\pm\)13 & 86\(\pm\)43 \\
\hline
\end{tabular}
\caption{The split and the description (mean and standard deviation) of the dataset. Para denotes paragraphs, sent denotes sentences, and punc denotes punctuations.}
\label{table:3}
\end{table*}

Linguistic features have been successfully applied to expose differences between deceptive and genuine content, so we subsume most of the features in previous works. The idea of explaining fictitious content is extended here to reveal how satirical news differs from true news. We divide our linguistic features into four families and compute them separately for paragraph and document. 

{\bf Psycholinguistic Features}: Psychological differences are useful for our problem, because professional journalists tend to express opinion conservatively to avoid unnecessary arguments. On the contrary, satirical news includes aggressive language for the entertainment purpose. We additionally observe true news favors clarity and accuracy while satirical news is related to emotional cognition. To capture the above observations, we employ Linguistic Inquiry and Word Count (LIWC)~\cite{pennebaker2007development} as our psycholinguistic dictionary. Each category of LIWC is one independent feature and valued by its frequency\footnote{Total counts divided by total words.}.

{\bf Writing Stylistic Features}: The relative distribution of part-of-speech (POS) tags reflects informative vs. imaginative writing, which contributes to detecting deceptions~\cite{li2014towards,mukherjee2013spotting}. We argue that the stories covered by satirical news are based on imagination. In addition, POS tags are hints of the underlying humor~\cite{reyes:2012humor}, which is common in satirical news. So we utilize POS tags~\cite{toutanova2003feature} to apprehend satire. Each tag is regarded as one independent feature and valued by its frequency.   

{\bf Readability Features}: We consider readability of genuine news would differ from satirical news because the former is written by professional journalists and tend to be clearer and more accurate, while satirical news packs numerous clauses to enrich the made-up story as introduced by~\newcite{rubin2016fake}. Different from their work, we use readability metrics, including Flesch Reading Ease~\cite{kincaid1975derivation}, Gunning Fog Index~\cite{gunning1952technique}, Automated Readability Index~\cite{senter1967automated}, ColemanLiau Index~\cite{coleman1975computer}, and syllable count per word, as features. 

{\bf Structural Features}: To further reflect the structure of news articles, we examine the following features: word count, log word count, number of punctuations, number of digits, number of capital letters, and number of sentences.

\section{Experiment and Evaluation}
We report satirical news detection results and show high weighted word features. Then, we provide a thorough analysis between paragraph-level and document-level features. Finally, we visualize an example of satirical news article to demonstrate the effectiveness of our work.

\subsection{Dataset}
The satirical news is collected from 14 websites that explicitly declare they are offering satire, so the correct label can be guaranteed. We also notice websites that mix true news, fake news, and satirical news. We exclude these websites in this work because it requires experts to annotate the news articles.

We maintain each satire source in only one of the train/validation/test sets\footnote{Train: Onion, the Spoof. Test: SatireWorld, Beaverton, Ossurworld. Validation: DailyCurrent, DailyReport, EnduringVision, Gomerblog, NationalReport, SatireTribune, SatireWire, Syruptrap, and UnconfirmedSource.} as the cross-domain setting in~\cite{li2014towards}. Otherwise, the problem may become writing pattern recognition or news site classification. We also combined different sources together\footnote{The combination is chosen to ensure enough training examples and balanced validation/test sets.} as a similar setting of leveraging multiple domains~\cite{yang2016leveraging}. The true news is collected from major news outlets\footnote{CNN, DailyMail, WashingtonPost, NYTimes, TheGuardian, and Fox.} and Google News using FLORIN~\cite{liu2015florin}. The satirical news in the corpus is significantly less than true news, reflecting an impressionistic view of the reality. We omit headline, creation time, and author information so this work concentrates on the satire in the article body. We realize the corpus may contain different degree of satire. Without the annotation, we only consider binary classification in this work and leave the degree estimation for the future. The split and the description of the dataset can be found in Table~\ref{table:3}. 

\begin{table*}[ht]
\centering
\small
\begin{tabular}[width=\textwidth]{l||*{4}{c|}*{4}{|c}}
\hline
\multicolumn{1}{c||}{Model} &\multicolumn{4}{c||}{Validation}&\multicolumn{4}{c}{Test} \\
\hline
& Acc & Pre & Rec & F1 & Acc & Pre & Rec & F1 \\
\hline
SVM word n-grams & 97.69 & 87.45 & 84.66 & 86.03 & 97.46 & 89.59 & 83.45 & 86.41\\
\hline
SVM word n-grams + LF & 97.73 & 86.06 & {\bf87.14} & 86.60 & 97.52 & 88.44 & 85.48 & 86.93\\
\hline
SVM word + char n-grams & 97.43 & 87.10 & 81.57 & 84.24 & 97.64 & 90.76 & 84.12 & 87.31\\
\hline
SVM word + char n-grams + LF & 97.76 & 90.13 & 82.44 & 86.11 & 97.93 & 92.71 & 85.31 & 88.86\\
\hline
SVM~\newcite{rubin2016fake} & 97.73 & 90.21 & 81.92 & 85.86 & 97.79 & {\bf93.47} & 82.95 & 87.90 \\
\hline
SVM~\newcite{rubin2016fake} + char tf-idf + LF & {\bf97.93} & {\bf90.99} & 83.69 & {\bf87.19} & {\bf98.09} & 92.98 & {\bf86.72} & {\bf89.75} \\
\hline
\hline
Bi-GRU & 97.67 & 89.17 & 82.28 & 85.58 & 97.58 & 93.11 & 80.96 & 86.61 \\
\hline
SVM Doc2Vec~\newcite{le:2014distributed} & 92.48 & 58.48 & 71.66 & 64.40 & 90.48 & 50.52 & 67.88 & 57.92 \\
\hline
HAN~\newcite{yang:2016hierarchical} & 97.91 & 92.06 & 82.24 & 86.88 & 97.83 & 90.85 & 86.17 & 88.45 \\
\hline
\hline
4LHN & 98.44 & 92.82 & 88.33 & 90.52 & 98.36 & 94.61 & 88.00 & 91.18\\
\hline
4LHNP & 98.46 & 93.54 & 87.75 & 90.56 & 98.39 & 94.63 & 88.33 & 91.37\\
\hline
4LHND & 98.36 & {\bf 94.73} & 85.24 & 89.74 & 98.18 & {\bf 95.35} & 85.31 & 90.05\\
\hline
4LHNPD & {\bf 98.54} & 93.31 & {\bf 89.01} & {\bf 91.11} & {\bf 98.39} & 93.51 & {\bf 89.50} & {\bf 91.46}\\
\hline
\end{tabular}
\caption{Satirical news detection results.}
\label{table:4}
\end{table*}

\subsection{Implementation Detail}
For SVM, we use the sklearn implementation\footnote{\href{http://scikit-learn.org/stable/modules/generated/sklearn.svm.SVC.html}{sklearn.svm.SVC}}. We find that using linear kernel and setting ``class\_weight" to ``balanced" mostly boost the result. We search soft-margin penalty ``C" and find high results occur in range~\([10^{-1},10^{-4}]\). 
We use the validation set to tune the model so selecting hyper-parameters is consistent with neural network based model. 

For neural network based models, we use the Theano package~\cite{Theano:2012} for implementation. The lengths of words, paragraphs, and documents are fixed at 24, 128, and 16 with necessary padding or truncating. Stochastic Gradient Descent is used with initial learning rate of 0.3 and decay rate of 0.9. The training is early stopped if the F1 drops 5 times continuously. Word embeddings are initialized with 100-dimension Glove embeddings~\cite{pennington2014glove}. Character embeddings are randomly initialized with 30 dimensions. Specifically for the proposed model, the following hyper-parameters are estimated based on the validation set and used in the final test set. The dropout is applied with probability of 0.5. The size of the hidden states is set at 60. We use 30 filters with window size of 3 for convolution. 


\subsection{Performance of Satirical News Detection}
We report accuracy, precision, recall, and F1 on the validation set and the test set. All metrics take satirical news as the positive class. Both paragraph-level and document-level linguistic features are scaled to have zero mean and unit variance, respectively. The compared methods include:

{\bf SVM word n-grams}: Unigram and bigrams of the words as the baseline. We report 1,2-grams because it performs better than other n-grams.


{\bf SVM word n-grams + LF}: 1,2-word grams plus linguistic features. We omit comparison with similar work~\cite{ott:2011finding} as their features are subsumed in ours.

{\bf SVM word + char n-grams}: 1,2-word grams plus bigrams and trigrams of the characters.

{\bf SVM word + char n-grams + LF}: All the proposed features are considered.

{\bf SVM \newcite{rubin2016fake}}: Unigram and bigrams tf-idf with satirical features as proposed in~\cite{rubin2016fake}. We compare with~\cite{rubin2016fake} rather than~\cite{burfoot:2009automatic} as the former claims a better result.

{\bf SVM \newcite{rubin2016fake} + char tf-idf + LF}: Include all possible features.

{\bf Bi-GRU}: Bi-GRU for document classification. The document representation is the average of the hidden state at every time-step.

{\bf SVM Doc2Vec}: Unsupervised method learning distributed representation for documents~\cite{le:2014distributed}. The implementation is based on Gensim~\cite{rehurek_lrec}.

{\bf HAN}: Hierarchical Attention Network~\cite{yang:2016hierarchical} for document classification with both word-level and sentence-level attention.

{\bf 4LHN}: 4-Level Hierarchical Network without any linguistic features.

{\bf 4LHNP}: 4-Level Hierarchical Network with Paragraph-level linguistic features.

{\bf 4LHND}: 4-Level Hierarchical Network with Document-level linguistic features.  

{\bf 4LHNPD}: 4-Level Hierarchical Network with both Paragraph-level and Document-level linguistic features. 

In Table~\ref{table:4}, the performances on the test set are generally better than on the validation set due to the cross-domain setting. We also explored word-level attention~\cite{yang:2016hierarchical}, but it performed 2\% worse than 4LHN. The result of Doc2Vec is limited. We suspect the reason could be the high imbalanced dataset, as an unsupervised learning method for document representation heavily relies on the distribution of the document.


\subsection{Word Level Analysis}
\label{wla}

\begin{table}[ht]
\centering
\small
\begin{tabular}[width=\columnwidth]{ll|ll}
\hline
\multicolumn{2}{c|}{\small True} & \multicolumn{2}{c}{\small Satire} \\
\hline
: & day & \textquotesingle\textquotesingle & stated  \\
video & said the & sources & press \\
but the & twitter & continued & reporter \\
in statement & told the & added & resident \\
com & pictured & washington dc & said that \\
\hline
\end{tabular}
\caption{High weighted word-level features}
\label{table:5}
\end{table}

\begin{table*}[ht]
\centering
\small
\centerline{
\begin{tabular}[width=\textwidth]{*{5}{@{\hspace{1.5mm}}c@{\hspace{1.5mm}}|}|*{5}{@{\hspace{1.5mm}}c@{\hspace{1.5mm}}|}*{5}{|@{\hspace{1.5mm}}c@{\hspace{1.5mm}}}}
\hline
\multicolumn{5}{c}{Psycholinguistic Feature}&\multicolumn{5}{c}{Writing Stylistic Feature}&\multicolumn{5}{c}{Readability Feature} \\
\hline
Name & S.m & S.std & T.m & T.std & Name & S.m & S.std & T.m & T.std & Name & S.m & S.std & T.m & T.std \\
\hline
{\bf Human.P} & .011 & .021 & .009 & .023 & {\bf JJ.P} & .061 & .045 & .058 & .046 & {\bf FRE.D} & 58.4 & 12.2 & 56.0 & 10.1 \\
\hline
{\bf Past.P}  & .034 & .035 & .040 & .042 & {\bf PRP.P} & .054 & .047 & .044 & .047 & {\bf CLI.D} & 9.08 & 1.66 & 9.48 & 1.61 \\
\hline
{\bf Self.P}  & .017 & .032 & .010 & .027 & {\bf RB.P} & .051 & .048 & .045 & .054 & {\bf FOG.D} & 13.71 & 3.25 & 14.00 & 2.89 \\
\hline 
{\bf Funct.D} & .453 & .045 & .437 & .049 & {\bf VBN.P} & .021 & .026 & .024 & .031 & \multicolumn{5}{c}{Structural Feature} \\ 
\hline
{\bf Social.P} & .097 & .067 & .091 & .073 & {\bf NN.D} & .273 & .038 & .300 & .043 & {\bf Punc.P} & 7.69 & 5.35 & 4.69 & 3.83\\
\hline
{\bf Leisure.P} & .017 & .027 & .018 & .032 & {\bf VBZ.P} & .019 & .026 & .021 & .029 & {\bf Cap.P} & 7.44 & 6.08 & 5.75 & 4.8 \\
\hline 
{\bf Hear.P} & .011 & .019 & .012 & .021 & {\bf CC.P} & .023 & .024 & .024 & .026 & {\bf Digit.P} & 0.97 & 2.40 & 1.39 & 3.00 \\
\hline
{\bf Bio.P} & .026 & .035 & .023 & .036 & {\bf CD.P} & .013 & .027 & .024 & .043 & {\bf LogWc.P} & 3.69 & 0.71 & 3.39 & 0.53 \\
\hline
\end{tabular}
}
\caption{\fontsize{10}{11}\selectfont Comparing feature values within each category. P stands for paragraph level. D stands for document level. S stands for satirical news. T stands for true news. m stands for mean and std stands for standard deviation. FRE: Flesch Reading Ease, the lower the harder. CLI: ColemanLiau Index. FOG: Gunning Fog Index. Punc: punctuation. Cap: Capital letters. LogWc: Log Word count}
\label{table:6}
\end{table*}

We report high weighted word-grams in Table~\ref{table:5} based on the SVM model as incorporating word-level attention in our neural hierarchy model reduces the detection performance. According to Table~\ref{table:5}, we conclude satirical news mimics true news by using news related words, such as ``stated" and ``reporter". However, these words may be over used so they can be detected. True news may use other evidence to support the credibility, which explains ``twitter", ``com", ``video", and ``pictured". High weight of `` : " indicates that true news uses colon to list items for clarity. High weight of `` \textquotesingle\textquotesingle \space" indicates that satirical news involves more conversation, which is consistent with our observation. The final interesting note is satirical news favors ``washington dc". We suspect that satirical news mostly covers politic topics, or satire writers do not spend efforts on changing locations.  

\subsection{Analysis of Weighted Linguistic Features}

We use 4LHNPD to compare paragraph-level and document-level features, as 4LHNPD leverages the two-level features into the same framework and yields the best result. 

Because all linguistic features are leveraged into
MLP with non-linear functions, it is hard to check which feature indicates satire. Alternatively, we define the importance of linguistic features by summing the absolute value of the weights if directly connected to the feature. For example, the importance~\(\mathrm{I}\) of feature~\(k\) is given by~\(\mathrm{I}_{k}=\frac{1}{M}\sum_{m=0}^M|\mathbf{w}_{k,m}|\), where~\(\mathbf{w}\in\mathbb{R}^{K\times M}\) is the directly connected weight,~\(K\) is the number of features, and~\(M\) is the dimension of the output. This metric gives a general idea about how much does a feature contribute to the decision making.  

We first report the scaled importance of the four linguistic feature sets by averaging the importance of individual linguistic features. Then we report individual important features within each set. 

\subsubsection{Comparing the Four Feature Sets}

\begin{figure}[ht]
\centering
\includegraphics[width=\columnwidth]{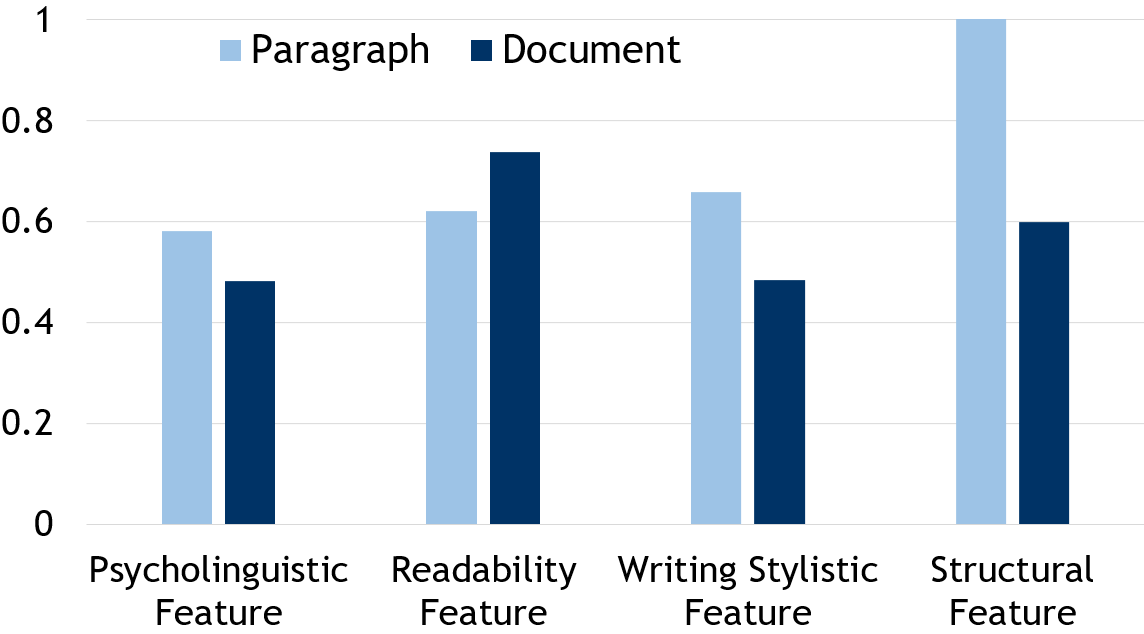}
\caption{Comparing the importance of the four feature sets at paragraph level and document level.}
\label{fig:2}
\end{figure}

According to Figure~\ref{fig:2}, the importance of paragraph-level features is greater than document-level features except for the readability feature set. It is reasonable to use readability at the document level because readability features evaluate the understandability of a given text, which depends on the content and the presentation. The structural feature set is highly weighted for selecting attended paragraph, which inspires us to focus on individual features inside the structural feature set.

\begin{figure*}[ht]
\centering
\includegraphics[width=\textwidth]{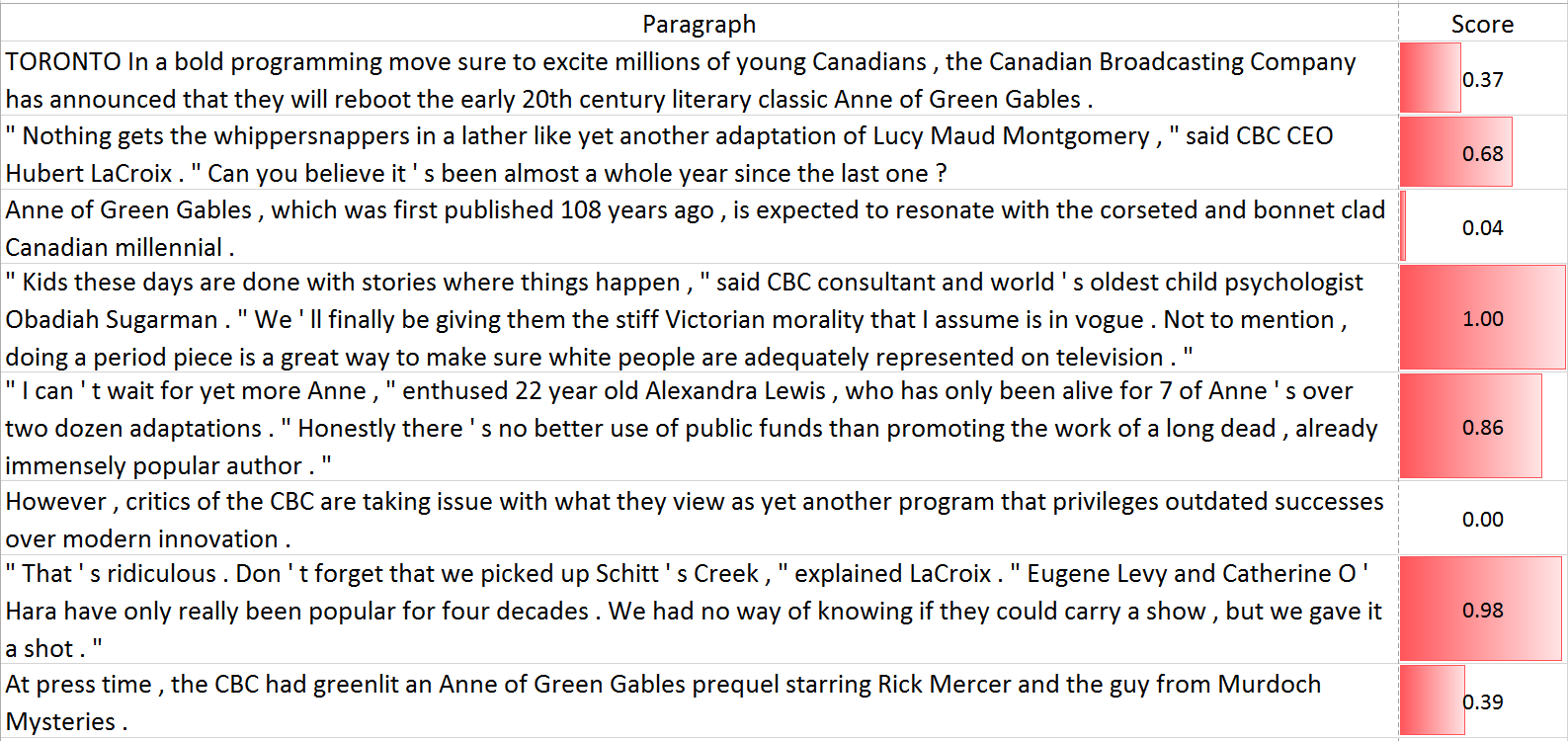}
\caption{An example of attended paragraphs.}
\label{fig:3}
\end{figure*}

\subsubsection{Comparing Individual Features}
Within each set, we rank features based on the importance score and report their mean and standard deviation before being scaled in Table~\ref{table:6}. At paragraph level, we use top three attended paragraphs for calculating. The respective p-values of all features in the table are less than 0.01 based on the t-test,
 indicating satirical news is statistically significantly different from true news.

Comparing Table~\ref{table:6} and Table~\ref{table:3}, we find that the word count, capital letters, and punctuations in true news are larger than in satirical news at the document level, while at paragraph level these features in true news are less than in satirical news. This indicates satire paragraph could be more complex locally. It also could be referred as ``sentence complexity", that {\it ``satirical articles tend to pack a great number of clauses into a sentence for comedic effect"}~\cite{rubin2016fake}. Accordingly, we hypothesize top complex paragraphs could represent the entire satire document for classification, which we leave for future examination.

In Table~\ref{table:6}, psycholinguistic feature ``Humans" is more related to emotional writing than control writing~\cite{pennebaker2007development}, which indicates satirical news is emotional and unprofessional compared to true news. The same reason also applies to ``Social" and ``Leisure", where the former implies emotional and the latter implies control writing. The ``Past" and ``VBN" both have higher frequencies in true news, which can be explained by the fact that true news covers what happened. A similar reason that true news reports what happened to others explains a low ``Self" and a high ``VBZ" in true news. 

For writing stylistic features, it is suggested that informative writing has more nouns, adjectives, prepositions and coordinating conjunctions, while imaginative writing has more verbs, adverbs, pronouns, and pre-determiners~\cite{rayson2001grammatical}. This explains higher frequencies of ``RB" and ``PRP" in satirical news, and higher frequency of ``NN" and ``CC" in true news. One exception is ``JJ", adjectives, which receives the highest weight in this feature set and indicates a higher frequency in satirical news. We suspect adjective could also be related to emotional writing, but more experiments are required.

Readability suggests satirical news is easier to be understood. Considering satirical news is also deceptive (as the story is not true), this is consistent with works~\cite{frank2008human,afroz:2012detecting} showing deceptive writings are more easily comprehended than genuine writings. Finally, true news has more digits and a higher ``CD"(Cardinal number) frequency, even at the paragraph level, because they tend to be clear and accurate.

\subsection{Visualization of Attended Paragraph}
To explore the attention, we sample one example in the validation set and present it in Figure~\ref{fig:3}.
The value at the right represents the scaled attention score. The high attended paragraphs are longer and have more capital letters as they are referring different entities. They have more double quotes, as multiple conversations are involved. 

Moreover,
we subjectively feel the attended paragraph with score 0.98 has a sense of humor while the paragraph with score 0.86 has a sense of sarcasm, which are common in satire. The paragraph with score 1.0 presents controversial topics, which could be misleading if the reader cannot understand the satire. This is what we expect from the attention mechanism. Based on the visualization, we also feel this work could be generalized to detect figurative languages.

\section{Conclusion}
In this paper, we proposed a 4-level hierarchical network and utilized attention mechanism to understand satire at both paragraph level and document level. The evaluation suggests readability features support the final classification while psycholinguistic features, writing stylistic features, and structural features are beneficial at the paragraph level. In addition, although satirical news is shorter than true news at the document level, we find satirical news generally contain paragraphs which are more complex than true news at the paragraph level. The analysis of individual features reveals that the writing of satirical news tends to be emotional and imaginative.

We will investigate efforts to model satire at the paragraph level following our conclusion and theoretical backgrounds, such as~\cite{ermida2012news}. We plan to go beyond the binary classification and explore satire degree estimation. We will generalize our approach to reveal characteristics of figurative language~\cite{joshi:2016automatic}, where different paragraphs or sentences may reflect different degrees of sarcasm, irony, and humor. 

\section*{Acknowledgments}

The authors would like to thank the anonymous reviewers for their comments. This work was support in part by the U.S. NSF grants 1546480 and 1527364.

\bibliography{emnlp2017}

\begin{thebibliography}{52}
\expandafter\ifx\csname natexlab\endcsname\relax\def\natexlab#1{#1}\fi

\bibitem[{Afroz et~al.(2012)Afroz, Brennan, and
  Greenstadt}]{afroz:2012detecting}
Sadia Afroz, Michael Brennan, and Rachel Greenstadt. 2012.
\newblock Detecting hoaxes, frauds, and deception in writing style online.
\newblock In \emph{2012 IEEE Symposium on Security and Privacy}, pages
  461--475. IEEE.

\bibitem[{Bahdanau et~al.(2014)Bahdanau, Cho, and Bengio}]{bahdanau:2014neural}
Dzmitry Bahdanau, Kyunghyun Cho, and Yoshua Bengio. 2014.
\newblock Neural machine translation by jointly learning to align and
  translate.
\newblock \emph{arXiv preprint arXiv:1409.0473}.

\bibitem[{Barbieri et~al.(2015)Barbieri, Ronzano, and Saggion}]{barbieri2015we}
Francesco Barbieri, Francesco Ronzano, and Horacio Saggion. 2015.
\newblock Do we criticise (and laugh) in the same way? automatic detection of
  multi-lingual satirical news in twitter.
\newblock In \emph{IJCAI}, pages 1215--1221.

\bibitem[{Bastien et~al.(2012)Bastien, Lamblin, Pascanu, Bergstra, Goodfellow,
  Bergeron, Bouchard, Warde-Farley, and Bengio}]{Theano:2012}
Fr{\'e}d{\'e}ric Bastien, Pascal Lamblin, Razvan Pascanu, James Bergstra, Ian
  Goodfellow, Arnaud Bergeron, Nicolas Bouchard, David Warde-Farley, and Yoshua
  Bengio. 2012.
\newblock Theano: new features and speed improvements.
\newblock \emph{arXiv preprint arXiv:1211.5590}.

\bibitem[{Burfoot and Baldwin(2009)}]{burfoot:2009automatic}
Clint Burfoot and Timothy Baldwin. 2009.
\newblock Automatic satire detection: Are you having a laugh?
\newblock In \emph{Proceedings of the ACL-IJCNLP 2009 conference short papers},
  pages 161--164. Association for Computational Linguistics.

\bibitem[{Castillo et~al.(2011)Castillo, Mendoza, and
  Poblete}]{castillo:2011information}
Carlos Castillo, Marcelo Mendoza, and Barbara Poblete. 2011.
\newblock Information credibility on twitter.
\newblock In \emph{Proceedings of the 20th international conference on World
  wide web}, pages 675--684. ACM.

\bibitem[{Chen et~al.(2016{\natexlab{a}})Chen, Bolton, and
  Manning}]{chen:2016thorough}
Danqi Chen, Jason Bolton, and Christopher~D Manning. 2016{\natexlab{a}}.
\newblock A thorough examination of the cnn/daily mail reading comprehension
  task.
\newblock \emph{arXiv preprint arXiv:1606.02858}.

\bibitem[{Chen et~al.(2016{\natexlab{b}})Chen, Sun, Tu, Lin, and
  Liu}]{chen2016neural}
Huimin Chen, Maosong Sun, Cunchao Tu, Yankai Lin, and Zhiyuan Liu.
  2016{\natexlab{b}}.
\newblock Neural sentiment classification with user and product attention.
\newblock In \emph{Proceedings of EMNLP}.

\bibitem[{Chen et~al.(2015)Chen, Conroy, and Rubin}]{chen:2015misleading}
Yimin Chen, Niall~J Conroy, and Victoria~L Rubin. 2015.
\newblock Misleading online content: Recognizing clickbait as false news.
\newblock In \emph{Proceedings of the 2015 ACM on Workshop on Multimodal
  Deception Detection}, pages 15--19. ACM.

\bibitem[{Cho et~al.(2014)Cho, Van~Merri{\"e}nboer, Gulcehre, Bahdanau,
  Bougares, Schwenk, and Bengio}]{cho2014learning}
Kyunghyun Cho, Bart Van~Merri{\"e}nboer, Caglar Gulcehre, Dzmitry Bahdanau,
  Fethi Bougares, Holger Schwenk, and Yoshua Bengio. 2014.
\newblock Learning phrase representations using rnn encoder-decoder for
  statistical machine translation.
\newblock \emph{arXiv preprint arXiv:1406.1078}.

\bibitem[{Coleman and Liau(1975)}]{coleman1975computer}
Meri Coleman and Ta~Lin Liau. 1975.
\newblock A computer readability formula designed for machine scoring.
\newblock \emph{Journal of Applied Psychology}, 60(2):283.

\bibitem[{Dong et~al.(2015)Dong, Gabrilovich, Murphy, Dang, Horn, Lugaresi,
  Sun, and Zhang}]{dong2015knowledge}
Xin~Luna Dong, Evgeniy Gabrilovich, Kevin Murphy, Van Dang, Wilko Horn, Camillo
  Lugaresi, Shaohua Sun, and Wei Zhang. 2015.
\newblock Knowledge-based trust: Estimating the trustworthiness of web sources.
\newblock \emph{Proceedings of the VLDB Endowment}, 8(9):938--949.

\bibitem[{Ermida(2012)}]{ermida2012news}
Isabel Ermida. 2012.
\newblock News satire in the press: Linguistic construction of humour inspoof
  news articles.
\newblock \emph{Language and humour in the media}, page 185.

\bibitem[{Farajtabar et~al.(2017)Farajtabar, Yang, Ye, Xu, Trivedi, Khalil, Li,
  Song, and Zha}]{farajtabar2017fake}
Mehrdad Farajtabar, Jiachen Yang, Xiaojing Ye, Huan Xu, Rakshit Trivedi, Elias
  Khalil, Shuang Li, Le~Song, and Hongyuan Zha. 2017.
\newblock Fake news mitigation via point process based intervention.
\newblock \emph{arXiv preprint arXiv:1703.07823}.

\bibitem[{Feng et~al.(2012)Feng, Banerjee, and Choi}]{feng:2012syntactic}
Song Feng, Ritwik Banerjee, and Yejin Choi. 2012.
\newblock Syntactic stylometry for deception detection.
\newblock In \emph{Proceedings of the 50th Annual Meeting of the Association
  for Computational Linguistics: Short Papers-Volume 2}, pages 171--175.
  Association for Computational Linguistics.

\bibitem[{Frank et~al.(2008)Frank, Menasco, and O'Sullivan}]{frank2008human}
Mark~G Frank, Melissa~A Menasco, and Maureen O'Sullivan. 2008.
\newblock Human behavior and deception detection.
\newblock \emph{Wiley Handbook of Science and Technology for Homeland
  Security}.

\bibitem[{Ge et~al.(2013)Ge, Gao, Li, and Zhang}]{ge2013multi}
Liang Ge, Jing Gao, Xiaoyi Li, and Aidong Zhang. 2013.
\newblock Multi-source deep learning for information trustworthiness
  estimation.
\newblock In \emph{Proceedings of the 19th ACM SIGKDD international conference
  on Knowledge discovery and data mining}, pages 766--774. ACM.

\bibitem[{Gunning(1952)}]{gunning1952technique}
Robert Gunning. 1952.
\newblock The technique of clear writing.

\bibitem[{Gupta et~al.(2012)Gupta, Zhao, and Han}]{gupta:2012evaluating}
Manish Gupta, Peixiang Zhao, and Jiawei Han. 2012.
\newblock Evaluating event credibility on twitter.
\newblock In \emph{SDM}, pages 153--164. SIAM.

\bibitem[{Hochreiter and Schmidhuber(1997)}]{hochreiter:1997long}
Sepp Hochreiter and J{\"u}rgen Schmidhuber. 1997.
\newblock Long short-term memory.
\newblock \emph{Neural computation}, 9(8):1735--1780.

\bibitem[{Horne and Adali(2017)}]{horne2017just}
Benjamin~D Horne and Sibel Adali. 2017.
\newblock This just in: Fake news packs a lot in title, uses simpler,
  repetitive content in text body, more similar to satire than real news.
\newblock \emph{arXiv preprint arXiv:1703.09398}.

\bibitem[{Joshi et~al.(2016)Joshi, Bhattacharyya, and
  Carman}]{joshi:2016automatic}
Aditya Joshi, Pushpak Bhattacharyya, and Mark~James Carman. 2016.
\newblock Automatic sarcasm detection: A survey.
\newblock \emph{arXiv preprint arXiv:1602.03426}.

\bibitem[{Kincaid et~al.(1975)Kincaid, Fishburne~Jr, Rogers, and
  Chissom}]{kincaid1975derivation}
J~Peter Kincaid, Robert~P Fishburne~Jr, Richard~L Rogers, and Brad~S Chissom.
  1975.
\newblock Derivation of new readability formulas (automated readability index,
  fog count and flesch reading ease formula) for navy enlisted personnel.
\newblock Technical report, DTIC Document.

\bibitem[{Le and Mikolov(2014)}]{le:2014distributed}
Quoc~V Le and Tomas Mikolov. 2014.
\newblock Distributed representations of sentences and documents.
\newblock In \emph{ICML}, volume~14, pages 1188--1196.

\bibitem[{Li et~al.(2014{\natexlab{a}})Li, Ott, Cardie, and
  Hovy}]{li2014towards}
Jiwei Li, Myle Ott, Claire Cardie, and Eduard~H Hovy. 2014{\natexlab{a}}.
\newblock Towards a general rule for identifying deceptive opinion spam.
\newblock In \emph{ACL (1)}, pages 1566--1576. Citeseer.

\bibitem[{Li et~al.(2014{\natexlab{b}})Li, Li, Gao, Zhao, Fan, and
  Han}]{li2014resolving}
Qi~Li, Yaliang Li, Jing Gao, Bo~Zhao, Wei Fan, and Jiawei Han.
  2014{\natexlab{b}}.
\newblock Resolving conflicts in heterogeneous data by truth discovery and
  source reliability estimation.
\newblock In \emph{Proceedings of the 2014 ACM SIGMOD international conference
  on Management of data}, pages 1187--1198. ACM.

\bibitem[{Li et~al.(2016)Li, Meng, and Clement}]{li2016verification}
Xian Li, Weiyi Meng, and Yu~Clement. 2016.
\newblock Verification of fact statements with multiple truthful alternatives.
\newblock In \emph{12th International Conference on Web Information Systems and
  Technologies}.

\bibitem[{Li et~al.(2015)Li, Li, Gao, Su, Zhao, Fan, and Han}]{li2015discovery}
Yaliang Li, Qi~Li, Jing Gao, Lu~Su, Bo~Zhao, Wei Fan, and Jiawei Han. 2015.
\newblock On the discovery of evolving truth.
\newblock In \emph{Proceedings of the 21th ACM SIGKDD International Conference
  on Knowledge Discovery and Data Mining}, pages 675--684. ACM.

\bibitem[{Liu et~al.(2015)Liu, Dragut, Mukherjee, and Meng}]{liu2015florin}
Qingyuan Liu, Eduard~C Dragut, Arjun Mukherjee, and Weiyi Meng. 2015.
\newblock Florin: a system to support (near) real-time applications on user
  generated content on daily news.
\newblock \emph{Proceedings of the VLDB Endowment}, 8(12):1944--1947.

\bibitem[{Ma and Hovy(2016)}]{ma2016end}
Xuezhe Ma and Eduard Hovy. 2016.
\newblock \href {http://www.aclweb.org/anthology/P16-1101} {End-to-end sequence
  labeling via bi-directional lstm-cnns-crf}.
\newblock In \emph{Proceedings of the 54th Annual Meeting of the Association
  for Computational Linguistics (Volume 1: Long Papers)}, pages 1064--1074,
  Berlin, Germany. Association for Computational Linguistics.

\bibitem[{Mihalcea and Strapparava(2009)}]{mihalcea2009lie}
Rada Mihalcea and Carlo Strapparava. 2009.
\newblock The lie detector: Explorations in the automatic recognition of
  deceptive language.
\newblock In \emph{Proceedings of the ACL-IJCNLP 2009 Conference Short Papers},
  pages 309--312. Association for Computational Linguistics.

\bibitem[{Mukherjee et~al.(2013{\natexlab{a}})Mukherjee, Kumar, Liu, Wang, Hsu,
  Castellanos, and Ghosh}]{mukherjee2013spotting}
Arjun Mukherjee, Abhinav Kumar, Bing Liu, Junhui Wang, Meichun Hsu, Malu
  Castellanos, and Riddhiman Ghosh. 2013{\natexlab{a}}.
\newblock Spotting opinion spammers using behavioral footprints.
\newblock In \emph{Proceedings of the 19th ACM SIGKDD international conference
  on Knowledge discovery and data mining}, pages 632--640. ACM.

\bibitem[{Mukherjee et~al.(2013{\natexlab{b}})Mukherjee, Venkataraman, Liu, and
  Glance}]{mukherjee:2013yelp}
Arjun Mukherjee, Vivek Venkataraman, Bing Liu, and Natalie~S Glance.
  2013{\natexlab{b}}.
\newblock What yelp fake review filter might be doing?
\newblock In \emph{ICWSM}.

\bibitem[{Mukherjee and Weikum(2015)}]{mukherjee:2015leveraging}
Subhabrata Mukherjee and Gerhard Weikum. 2015.
\newblock Leveraging joint interactions for credibility analysis in news
  communities.
\newblock In \emph{Proceedings of the 24th ACM International on Conference on
  Information and Knowledge Management}, pages 353--362. ACM.

\bibitem[{Ott et~al.(2011)Ott, Choi, Cardie, and Hancock}]{ott:2011finding}
Myle Ott, Yejin Choi, Claire Cardie, and Jeffrey~T Hancock. 2011.
\newblock Finding deceptive opinion spam by any stretch of the imagination.
\newblock In \emph{Proceedings of the 49th Annual Meeting of the Association
  for Computational Linguistics: Human Language Technologies-Volume 1}, pages
  309--319. Association for Computational Linguistics.

\bibitem[{Pennebaker et~al.(2007)Pennebaker, Chung, Ireland, Gonzales, and
  Booth}]{pennebaker2007development}
James~W Pennebaker, Cindy~K Chung, Molly Ireland, Amy Gonzales, and Roger~J
  Booth. 2007.
\newblock The development and psychometric properties of liwc2007. austin, tx,
  liwc. net.

\bibitem[{Pennington et~al.(2014)Pennington, Socher, and
  Manning}]{pennington2014glove}
Jeffrey Pennington, Richard Socher, and Christopher~D Manning. 2014.
\newblock Glove: Global vectors for word representation.
\newblock In \emph{EMNLP}, volume~14, pages 1532--1543.

\bibitem[{P{\'e}rez-Rosas and Mihalcea(2014)}]{perez2014cross}
Ver{\'o}nica P{\'e}rez-Rosas and Rada Mihalcea. 2014.
\newblock Cross-cultural deception detection.
\newblock In \emph{ACL (2)}, pages 440--445.

\bibitem[{Potthast et~al.(2017)Potthast, Kiesel, Reinartz, Bevendorff, and
  Stein}]{potthast2017stylometric}
Martin Potthast, Johannes Kiesel, Kevin Reinartz, Janek Bevendorff, and Benno
  Stein. 2017.
\newblock A stylometric inquiry into hyperpartisan and fake news.
\newblock \emph{arXiv preprint arXiv:1702.05638}.

\bibitem[{Rayson et~al.(2001)Rayson, Wilson, and Leech}]{rayson2001grammatical}
Paul Rayson, Andrew Wilson, and Geoffrey Leech. 2001.
\newblock Grammatical word class variation within the british national corpus
  sampler.
\newblock \emph{Language and Computers}, 36(1):295--306.

\bibitem[{{\v R}eh{\r u}{\v r}ek and Sojka(2010)}]{rehurek_lrec}
Radim {\v R}eh{\r u}{\v r}ek and Petr Sojka. 2010.
\newblock {Software Framework for Topic Modelling with Large Corpora}.
\newblock In \emph{{Proceedings of the LREC 2010 Workshop on New Challenges for
  NLP Frameworks}}, pages 45--50, Valletta, Malta. ELRA.
\newblock \url{http://is.muni.cz/publication/884893/en}.

\bibitem[{Reyes et~al.(2012)Reyes, Rosso, and Buscaldi}]{reyes:2012humor}
Antonio Reyes, Paolo Rosso, and Davide Buscaldi. 2012.
\newblock From humor recognition to irony detection: The figurative language of
  social media.
\newblock \emph{Data \& Knowledge Engineering}, 74:1--12.

\bibitem[{Rockt{\"a}schel et~al.(2015)Rockt{\"a}schel, Grefenstette, Hermann,
  Ko{\v{c}}isk{\`y}, and Blunsom}]{rocktaschel2015reasoning}
Tim Rockt{\"a}schel, Edward Grefenstette, Karl~Moritz Hermann, Tom{\'a}{\v{s}}
  Ko{\v{c}}isk{\`y}, and Phil Blunsom. 2015.
\newblock Reasoning about entailment with neural attention.
\newblock \emph{arXiv preprint arXiv:1509.06664}.

\bibitem[{Rubin et~al.(2016)Rubin, Conroy, Chen, and Cornwell}]{rubin2016fake}
Victoria Rubin, Niall Conroy, Yimin Chen, and Sarah Cornwell. 2016.
\newblock \href {http://www.aclweb.org/anthology/W16-0802} {Fake news or truth?
  using satirical cues to detect potentially misleading news}.
\newblock In \emph{Proceedings of the Second Workshop on Computational
  Approaches to Deception Detection}, pages 7--17, San Diego, California.
  Association for Computational Linguistics.

\bibitem[{Senter and Smith(1967)}]{senter1967automated}
RJ~Senter and Edgar~A Smith. 1967.
\newblock Automated readability index.
\newblock Technical report, DTIC Document.

\bibitem[{Toutanova et~al.(2003)Toutanova, Klein, Manning, and
  Singer}]{toutanova2003feature}
Kristina Toutanova, Dan Klein, Christopher~D Manning, and Yoram Singer. 2003.
\newblock Feature-rich part-of-speech tagging with a cyclic dependency network.
\newblock In \emph{Proceedings of the 2003 Conference of the North American
  Chapter of the Association for Computational Linguistics on Human Language
  Technology-Volume 1}, pages 173--180. Association for Computational
  Linguistics.

\bibitem[{Wan et~al.(2016)Wan, Chen, Kaplan, Han, Gao, and Zhao}]{wan2016truth}
Mengting Wan, Xiangyu Chen, Lance Kaplan, Jiawei Han, Jing Gao, and Bo~Zhao.
  2016.
\newblock From truth discovery to trustworthy opinion discovery: An
  uncertainty-aware quantitative modeling approach.
\newblock In \emph{Proceedings of the 22nd ACM SIGKDD International Conference
  on Knowledge Discovery and Data Mining}, pages 1885--1894. ACM.

\bibitem[{Xiao et~al.(2016)Xiao, Gao, Li, Ma, Su, Feng, and
  Zhang}]{xiao2016towards}
Houping Xiao, Jing Gao, Qi~Li, Fenglong Ma, Lu~Su, Yunlong Feng, and Aidong
  Zhang. 2016.
\newblock Towards confidence in the truth: A bootstrapping based truth
  discovery approach.
\newblock In \emph{Proceedings of the 22nd ACM SIGKDD International Conference
  on Knowledge Discovery and Data Mining}, pages 1935--1944. ACM.

\bibitem[{Yang et~al.(2016{\natexlab{a}})Yang, Mukherjee, and
  Zhang}]{yang2016leveraging}
Fan Yang, Arjun Mukherjee, and Yifan Zhang. 2016{\natexlab{a}}.
\newblock Leveraging multiple domains for sentiment classification.
\newblock In \emph{Proceedings of COLING 2016, the 26th International
  Conference on Computational Linguistics: Technical Papers}, pages 2978--2988,
  Osaka, Japan. The COLING 2016 Organizing Committee.

\bibitem[{Yang et~al.(2016{\natexlab{b}})Yang, Yang, Dyer, He, Smola, and
  Hovy}]{yang:2016hierarchical}
Zichao Yang, Diyi Yang, Chris Dyer, Xiaodong He, Alex Smola, and Eduard Hovy.
  2016{\natexlab{b}}.
\newblock Hierarchical attention networks for document classification.
\newblock In \emph{Proceedings of the 2016 Conference of the North American
  Chapter of the Association for Computational Linguistics: Human Language
  Technologies}.

\bibitem[{Yin et~al.(2008)Yin, Han, and Philip}]{yin2008truth}
Xiaoxin Yin, Jiawei Han, and S~Yu Philip. 2008.
\newblock Truth discovery with multiple conflicting information providers on
  the web.
\newblock \emph{IEEE Transactions on Knowledge and Data Engineering},
  20(6):796--808.

\bibitem[{Zhao et~al.(2015)Zhao, Resnick, and Mei}]{zhao2015enquiring}
Zhe Zhao, Paul Resnick, and Qiaozhu Mei. 2015.
\newblock Enquiring minds: Early detection of rumors in social media from
  enquiry posts.
\newblock In \emph{Proceedings of the 24th International Conference on World
  Wide Web}, pages 1395--1405. ACM.

\end{thebibliography}
\bibliographystyle{emnlp_natbib}

\end{document}